\documentclass[10pt,twocolumn,letterpaper]{article}

\usepackage{iccv}
\usepackage{times}
\usepackage{epsfig}
\usepackage{graphicx}
\usepackage{amsmath}
\usepackage{amssymb}

\usepackage[pagebackref=true,breaklinks=true,letterpaper=true,colorlinks,bookmarks=false]{hyperref}
\usepackage[table,xcdraw]{xcolor}
\iccvfinalcopy 

\def\iccvPaperID{735} 

\ificcvfinal\fi
\begin{document}

\title{Identity-Aware Textual-Visual Matching with Latent Co-attention}

\author{Shuang Li, Tong Xiao, Hongsheng Li$^{*}$, Wei Yang, and Xiaogang Wang\thanks{Corresponding authors}\\
Department of Electronic Engineering, The Chinese University of Hong Kong\\
{\tt\small \{sli,xiaotong,hsli,wyang,xgwang\}@ee.cuhk.edu.hk}
}

\maketitle
\thispagestyle{empty}

\begin{abstract}
Textual-visual matching aims at measuring similarities between sentence descriptions and images. Most existing methods tackle this problem without effectively utilizing identity-level annotations.
In this paper, we propose an identity-aware two-stage framework for the textual-visual matching problem. Our stage-1 CNN-LSTM network learns to embed cross-modal features with a novel Cross-Modal Cross-Entropy (CMCE) loss.  The stage-1 network is able to efficiently screen easy incorrect matchings and also provide initial training point for the stage-2 training.
The stage-2 CNN-LSTM network refines the matching results with a latent co-attention mechanism.
The spatial attention relates each word with corresponding image regions while the latent semantic attention aligns different sentence structures to make the matching results more robust to sentence structure variations.
Extensive experiments on three datasets with identity-level annotations show that our framework outperforms state-of-the-art approaches by large margins.
\vspace{-20pt}
\end{abstract}

\vspace{-5pt}
\section{Introduction}
Identifying correspondences and measuring similarities between natural language descriptions and images is an important task in computer vision and has many applications, including text-image embedding~\cite{mao2014explain, Yan_2015_CVPR, wang2016learning, reed2016learning,livip}, zero-shot learning~\cite{palatucci2009zero, rohrbach2011evaluating, frome2013devise}, and visual QA~\cite{Antol_2015_ICCV, fukui2016multimodal, zhu2016visual7w, nam2016dual, lu2016hierarchical}. We call such a general problem \emph{textual-visual matching}, which has drawn increasing attention in recent years. The task is challenging because the complex relations between language descriptions and image appearance are highly non-linear and there exist large variations or subtle variations in image appearance for similar language descriptions.

There have been large scale image-language datasets and deep learning techniques~\cite{lin2014microsoft, krishna2016visual, Antol_2015_ICCV, young2014image, hodosh2013framing} proposed for textual-visual matching, which considerably advanced research progress along this direction. However, identity-level annotations provided in benchmark datasets are ignored by most existing methods when performing matching across textual and visual domains.

\begin{figure}
\begin{center}
\includegraphics[width=1\linewidth]{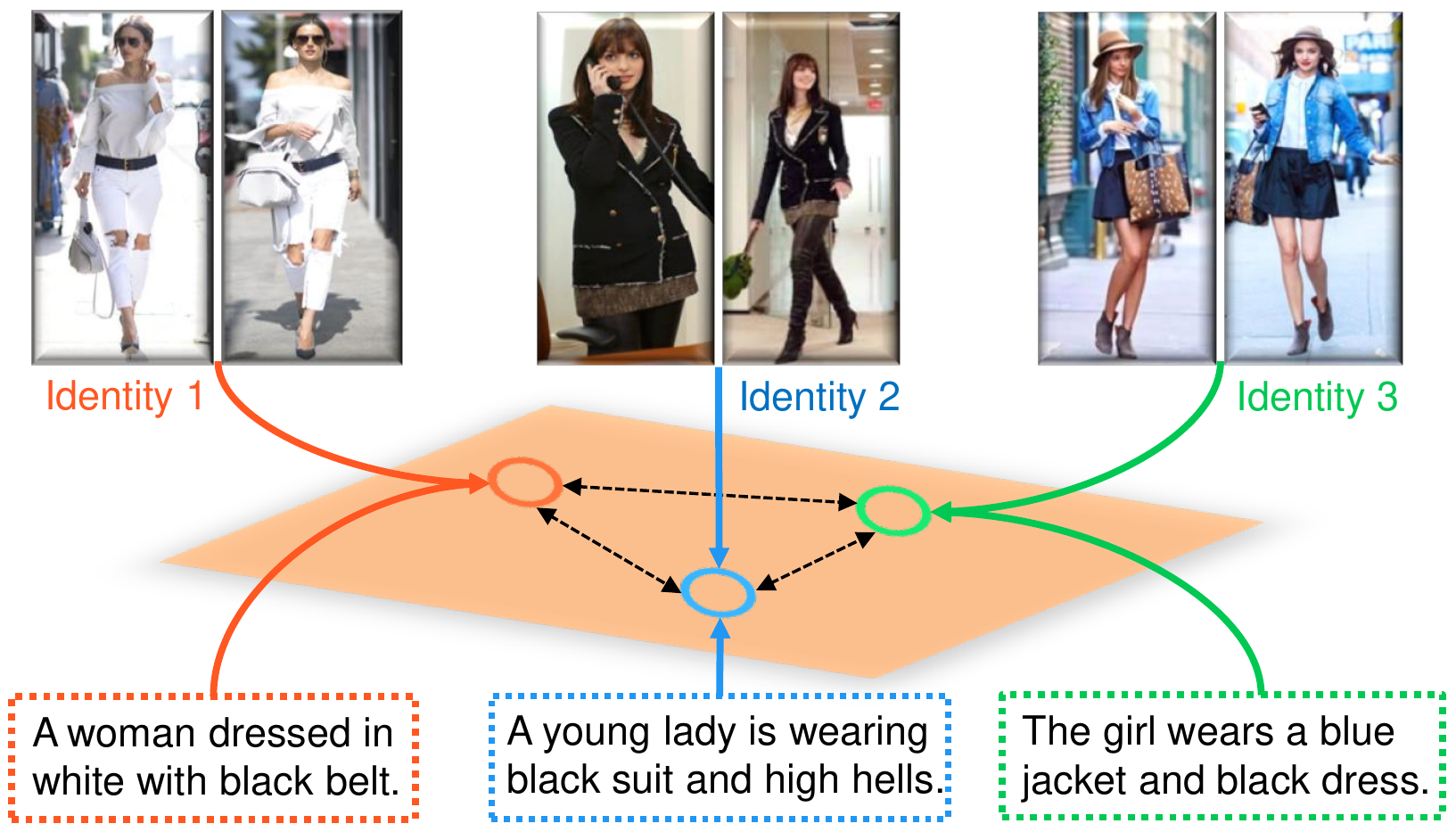} \ \\
\end{center}
\vspace{-8pt}
\caption{Learning deep features for textual-visual matching with identity-level annotations. Utilizing identity-level annotations could jointly minimize intra-identity discrepancy and maximize inter-identity discrepancy, and thus results in more discriminative feature representations.}
\label{fig:intro}
\vspace{-7pt}
\end{figure}

In this paper, we propose a two-stage framework for identity-aware textual-visual matching, which consists of two deep neural networks. The stage-1 network learns identity-aware feature representations of images and language descriptions by introducing a Cross-Modal Cross-Entropy (CMCE) loss to effectively utilize identity-level annotations for feature learning (see Figure \ref{fig:intro}). After training, it provides initial matching results and also serves as the initial point for training stage-2 network.
The stage-2 deep neural network employs a latent co-attention mechanism that jointly learns the spatial attention and latent semantic attention to match salient image regions and latent semantic concepts for textual-visual affinity estimation.

Our stage-1 network consists of a CNN and a LSTM for learning textual and visual feature representations. The objective is to minimize the feature distances between descriptions and images belonging to the same identities. 
The stage-1 network utilizes a specialized CMCE loss with dynamic buffers, which implicitly minimizes intra-identity feature distances and maximize inter-identity feature distances over the entire dataset instead of just small mini-batches. In contrast, for the pairwise or triplet loss functions, the probability of sampling hard negative samples during training decreases quadratically or cubically as the number of training sample increases.


The trained stage-1 network is able to efficiently screen easy incorrect matchings for both training and testing. However, one limitation of the CMCE loss in stage-1 is that the generated textual and visual features are not tightly coupled.
A further refinement on stage-1 results is essential for obtaining accurate matching results. Our stage-2 network is a tightly coupled CNN-LSTM network with latent co-attention. It takes a pair of language description and image as input and outputs the textual-visual matching confidence, which is trained with the binary cross-entropy loss.

Conventional RNNs for language encoding have difficulty in remembering the complete sequential information when the input descriptions are too long. It tends to miss important words appearing in the beginning of the sentence. 
The RNN is also variant to different sentence structures. Sentences describing the same image but with different sentence structures could be represented by features with large differences. For instance, ``the girl who has blond hair is wearing a white dress and heels'' and ``The girl wears heels and a white dress. She has blond hair.'' Both sentences describe the same person but the first one might focus more on ``white dress and heels'', and the second one might assign ``blond hair'' with higher weights.
Inspired by the word alignment (attention) technique in neural machine translation \cite{bahdanau2014neural}, a latent co-attention mechanism is proposed for the stage-2 CNN-LSTM network. The visual spatial attention module associates word to its related image regions. The latent semantic attention module aligns different sentence structures with an alignment decoder LSTM. At each step of the LSTM, it learns how to weight different words' features to be more invariant to sentence structure variations.


The contribution of this paper is three-fold. 1) We propose a novel identity-aware two-stage deep learning framework for solving the problem of textual-visual matching. The stage-1 network can efficiently screen easy incorrect matchings and also acts as the initial point for training stage-2 network. The stage-2 network refines matching results with binary classification. Identity-level annotations ignored by most existing methods are utilized to learn better feature representations. 2) To take advantage of the identity-level annotations, our stage-1 network employs a specialized CMCE loss with feature buffers. Such a loss enables effective feature embedding and fast evaluation. 3) A novel latent co-attention mechanism is proposed for our stage-2 network. It has a spatial attention module that focuses on relevant image regions for each input word, and a latent semantic attention module that automatically aligns different words' feature representations to minimize the impact of sentence structure variations.

\section{Related Work}

\subsection{Visual matching with identity-level annotations}
Visual matching tasks with identity-level annotations, such as person re-identification \cite{zhang2016learning,xiao2017joint,ahmed2015improved,liao2015person,zheng2016person,xiao2017joint,Zhao_2017_CVPR} and face recognition \cite{masi2016pose,schroff2015facenet}, are well-developed research areas. Visual matching algorithms either classify all the identities simultaneously \cite{liao2015person,kumar2009attribute,xiao2016learning} or learn pair-wise or triplet distance loss function \cite{ahmed2015improved,masi2016pose,schroff2015facenet,cheng2016person} for feature embedding. However, both of them have major limitations. The first type of loss function faces challenges  when the number of classes is too large. The limited number of classes (identities) in each mini-batch leads to unstable training behavior. For the second type of loss function, the hard negative training samples might be difficult to sample as the number of training sample increases, and the computation time of constructing pairs or triplets increases quadratically or cubically with the number of test samples.

\subsection{Textual-visual matching}
Measuring similarities between images and languages aims at understanding the relations between images and language descriptions. It gains a lot of attention in recent years because of its wide applications in image captioning \cite{mao2014deep,vinyals2015show,karpathy2015deep,Chen_2015_CVPR}, visual QA~\cite{Antol_2015_ICCV, zhu2016visual7w, nam2016dual, lu2016hierarchical}, and text-image embedding \cite{frome2013devise,reed2016learning,klein2015associating,Yan_2015_CVPR,wang2016learning}. Karpathy \etal \cite{karpathy2015deep} combined the convolutional neural network for image regions and bidirectional recurrent neural networks for descriptions to generate image captions.
 The word-image pairwise affinities are calculated for sentence-image ranking. Nam \etal \cite{nam2016dual} jointly learned image and language attention models to capture the shared concepts between the two domains and evaluated the affinity by computing the inner product of two fixed embedding vectors. \cite{Yan_2015_CVPR} tackled the matching problem with deep canonical correlation analysis by constructing the trace norm objective between image and language features. In \cite{klein2015associating}, Klein \etal presented two mixture models, Laplacian mixture model and Hybird Gaussian-Laplacian mixture model to learn Fisher vector representations of sentences. The text-to-image matching is conducted by associating the generated Fisher vector and VGG image features.

\subsection{Identity-aware textual-visual matching}
Although identity-level annotations are widely used in visual matching tasks, they are seldom exploited for textual-visual matching. Using such annotations can assist cross-domain feature embedding by minimizing the intra-identity distances and capturing the relations between textual concepts and visual regions, which makes textual-visual matching methods more robust to variations within each domain.

Reed \etal \cite{reed2016learning} collected fine-grained language descriptions for two visual datasets, Caltech-UCSD birds (CUB) and Oxford-102 Flowers, and first used the identity-level annotations for text-image feature learning. In \cite{li2017person}, Li \etal proposed a large scale person re-identification dataset with language descriptions and performed description-person image matching using an CNN-LSTM network with neural attention mechanism.
However, these approaches face the same problems with existing visual matching methods.
To solve these problems and efficiently learn textual and visual feature representations, we propose a novel two-stage framework for identity-aware textual-visual matching. Our approach outperforms both above state-of-the-art methods by large margins on the three datasets.

\vspace{-4pt}
\section{Identity-Aware Textual-Visual Matching with Latent Co-attention}
\vspace{-4pt}
Textual-visual matching aims at conducting accurate verification for images and language descriptions. However, identity-level annotations provided by many existing textual-visual matching datasets are not effectively exploited for cross-domain feature learning.
In this section, we introduce a novel identity-aware two-stage deep learning framework for textual-visual matching. The stage-1 CNN-LSTM network adopts a specialized Cross-Modal Cross-Entropy (CMCE) loss, which utilizes identity-level annotations to minimize intra-identity and maximize inter-identity feature distances. It is also efficient for evaluation because of its linear evaluation time. After training convergence, the stage-1 network is able to screen easy incorrect matchings and also provides initial point for training the stage-2 CNN-LSTM network. The stage-2 network further verifies hard matchings with a novel latent co-attention mechanism. It jointly learns the visual spatial attention and latent semantic attention in an end-to-end manner to recover the relations between visual regions and achieves robustness against sentence structure variations.

\subsection{Stage-1 CNN-LSTM with CMCE loss}
\begin{figure}
\begin{center}
\includegraphics[width=1\linewidth]{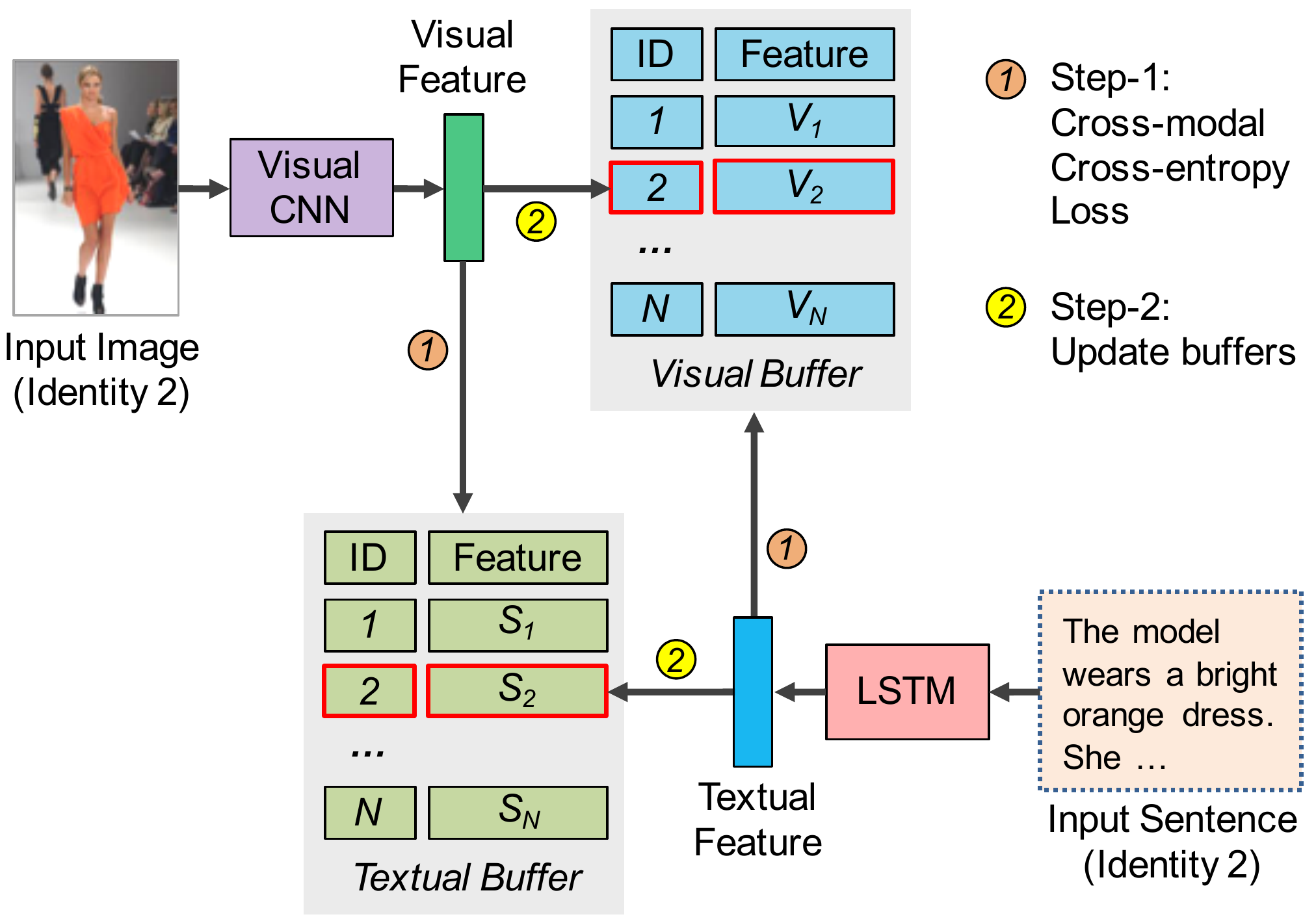} \ \\
\end{center}
\vspace{-9pt}
\caption{Illustration of the stage-1 network. In each iteration, the images and text descriptions in a mini-batch are first fed into the CNN and LSTM respectively to generate their feature representations. The CMCE loss is then computed by comparing sampled features in one modality to all other features in the feature buffer of the other modality (Step-1). The CNN and LSTM parameters are updated by backpropagation. Finally, the visual and textual feature buffers are updated with the sampled features (Step-2).}
\label{fig:stage1}
\end{figure}

\vspace{-4pt}
The structure of stage-1 network is illustrated in Figure \ref{fig:stage1}, which is a loosely coupled CNN-LSTM . Given an input textual description or image, both the visual CNN and language LSTM are trained to map the input image and description into a joint feature embedding space, such that the features representations belonging to the same identity should have small feature distances, while those of different identities should have large distances. To achieve the goal, the CNN-LSTM network is trained with a CMCE loss.

\vspace{-9pt}
\subsubsection{Cross-Modal Cross-Entropy Loss}
For the conventional pairwise classification loss \cite{Antol_2015_ICCV,lu2016hierarchical} or triplet max-margin loss \cite{wang2016learning,reed2016learning}, if there are $N$ identities in the training set, the number of possible training samples would be $O(N^2)$. It is generally difficult to sample hard negative samples for learning effective feature representations.
On the other hand, during evaluation phase, the time complexity of feature calculation of pairwise or triplet loss would increase quadratically with $N$, which would take lots of computation time.
To solve this problem, we propose a novel CMCE loss that efficiently compares a mini-batch of $n$ identity features from one modality to those of all $N$ identities in another modality in each iteration.
Intuitively, the sampled $n$ identity features are required to have high affinities with their corresponding identities in the other modality and low affinities with all other $N-n$ ones in the entire identity set. The cross-modal affinity is calculated as the inner products of features from the two modalities.
By using the proposed loss function, hard negative samples are all covered in each training epoch and the evaluation time complexity of sampling all test samples is only $O(N)$.

In each training iteration, a mini-batch of images belonging to $n$ different identities are transformed to visual features, each of which is denoted by $v \in \mathbb{R}^{D}$. $D$ is the feature embedding dimension for both modalities. Textual features of all $N$ identities are pre-stored in a textual feature buffer $S\in \mathbb{R}^{D\times N}$, where $S_i$ denotes the textual feature of the $i$th identity. The affinities between a visual feature representation $v$ and all textual features $S$ could then be calculated as $S^T v$.
The probability of the input image $v$ matching the $i$th identity in the textual feature buffer can be calculated with the following cross-modal softmax function,
\begin{align}
p_{i}^{S}(v)=\dfrac{\exp{(S_i^T v/\sigma_v)}}{\sum_{j=1}^{N}{\exp{(S_j^T v/\sigma_v)}}},
\label{eq:1}
\end{align}
where $\sigma_{v}$ is a temperature hyper-parameter to control how peaky the probability distribution is.
Similarly, in each iteration, a mini-batch of sentence descriptions belonging to $n$ identities are also sampled. Let $s \in \mathbb{R}^D$ denote one text sample's feature in the mini-batch. All visual features are pre-stored in a visual feature buffer $V\in \mathbb{R}^{D\times N}$. The probability of $s$ matching the $k$th identity in the visual feature buffer is defined as
\begin{align}
p_{k}^{V}(s)=\dfrac{\exp{(V_k^T s/\sigma_s)}}{\sum_{j=1}^{N}{\exp{(V_j^T s/\sigma_s)}}},
\label{eq:2}
\end{align}
where $\sigma_{s}$ is another temperature hyper-parameter.
In each iteration, our goal is to maximize the above textual and visual matching probabilities for corresponding identity pairs. The learning objective can then be define as minimizing the following CMCE loss,
\begin{align}
L=-\sum_{v}{\log{p_{t_v}^{S}} (v)}-\sum_{s}{\log{p_{t_s}^{V}} (s)},
\end{align}
where $t_v$ and $t_s$ are the target identities of visual feature $v$ and textual feature $s$ respectively. Its gradients are calculated as
\begin{align}
\dfrac{\partial{L}}{\partial{v}} = \dfrac{1}{\sigma_v} \left[ (p_{t_v}^{S}-1) S_{t_v}+ \sum_{\substack{j=1\\j\neq{t_v}}}^{N}{S_j p_{j}^{S}} \right],\\
\dfrac{\partial{L}}{\partial{s}}=\dfrac{1}{\sigma_s} \left[ (p_{t_s}^{V}-1) V_{t_s} + \sum_{\substack{j=1\\j\neq{t_s}}}^{N}{V_j p_{j}^{V}} \right].
\end{align}

The textual and visual feature buffers enable efficient calculation of textual-visual affinities between sampled identity features in one modality and all features in the other modality. This is the key to our cross-modal entropy loss.
Before the first iteration, image and textual features are obtained by the CNN and LSTM. Each identity's textual and visual features are stored in its corresponding row in the textual and visual feature buffers. If an identity has multiple descriptions or images, its stored features in the buffers are the average of the multiple samples. In each iteration, after the forward propagation, the loss function is first calculated. The parameters of both visual CNN and language LSTM are updated via backpropgation. For the sampled identity images and descriptions, their corresponding rows in the textual and visual feature buffers are updated by the newly generated features. If a corresponding identity $t$ has multiple entity images or descriptions, the buffer rows are updated as the running weighted average with the following formulations, $S_{t_v} = 0.5 S_{t_v} + 0.5 s$ and $V_{t_s} = 0.5 V_{t_s} + 0.5 v$, where $s$ and $v$ are the newly generated textual and visual features, $t_s$ and $t_v$ denote their corresponding identities.

Although our CMCE loss has similar formation with softmax loss function, they have major differences. First, the CMCE propagates gradients across textual and visual domains. It can efficiently embed features of the same identity from different domains to be similar and enlarge the distances between non-corresponding identities. Second, the feature buffers store all identities' feature representations of different modalities, making the comparison between mini-batch samples with all identities much efficient.

\subsection{Stage-2 CNN-LSTM with latent co-attention}
\begin{figure}[t]
\begin{center}
\includegraphics[width=1\linewidth]{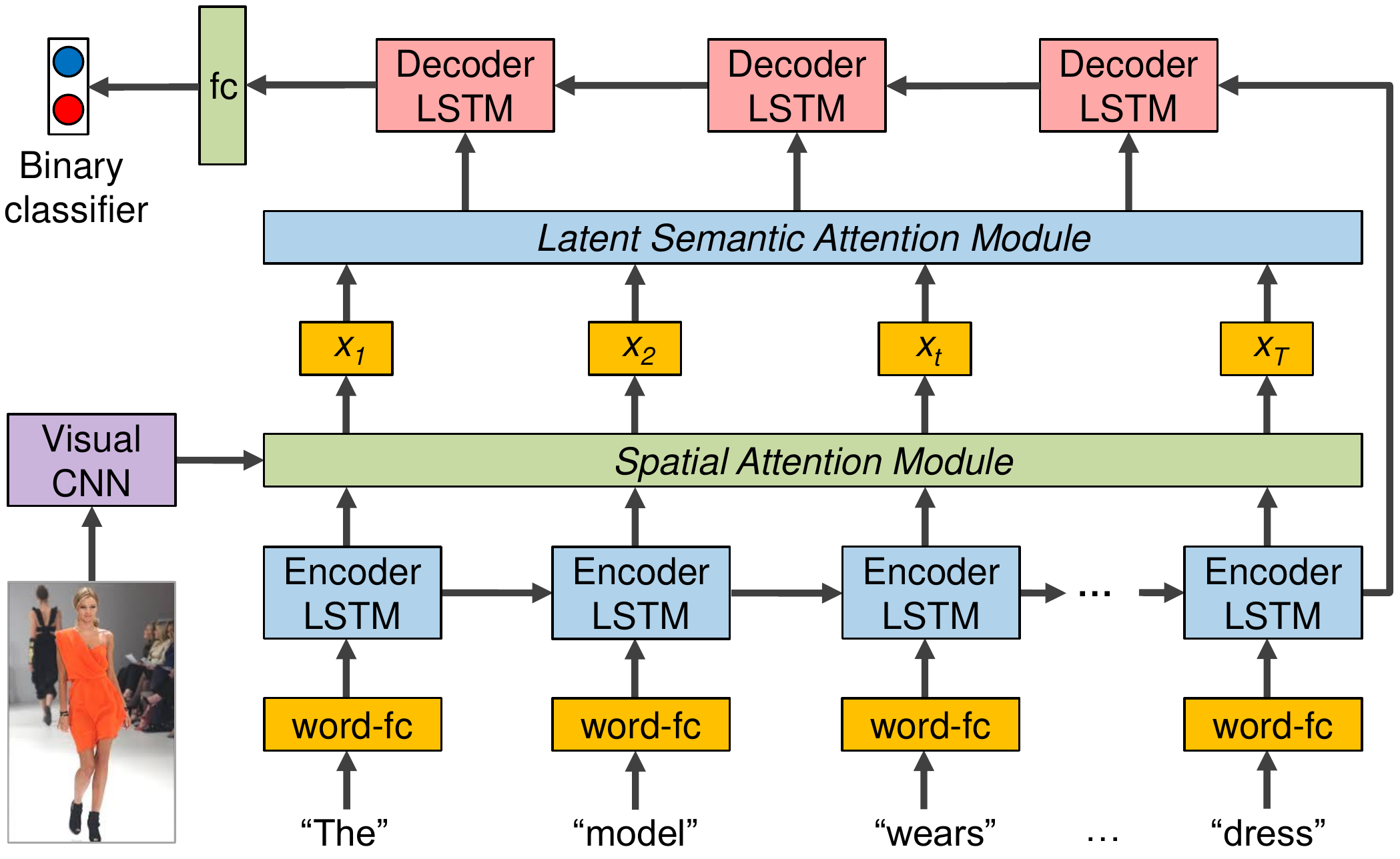} \ \\
\end{center}
\vspace{-7pt}
\caption{Illustration of the stage-2 network with latent co-attention mechanism. The spatial attention associates the relevant visual regions to each input word while the latent semantic attention automatically aligns image-word features by the spatial attention modules to enhance the robustness to sentence structure variations.}
\label{fig:stage2}
\end{figure}
After training, the stage-1 network is able to obtain initial matching results efficiently because the textual and visual features can be calculated independently for each modality.
However, the visual and text feature embeddings might not be optimal because stage-1 compresses the whole sentence into a single vector. The correspondences between individual words and image regions are not established to capture word-level similarities. Stage-1 is also sensitive to sentence structure variations.
A further refinement on the stage-1 matching results is desirable for obtaining accurate matching results. For stage 2, we propose a tightly coupled CNN-LSTM network with latent co-attention mechanism, which takes a pair of text description and image as inputs and outputs their matching confidence.
Stage-2 framework associates individual words and image regions with spatial attention to better capture world-level similarities, and automatically realigns sentence structures via latent semantic attention.
The trained stage-1 network serves as the initial point for the stage-2 network. In addition, it screens easy negatives, so only the hard negative matching samples from stage-1 results are utilized for training stage-2.
With stage-1, stage-2 can focus on handling more challenging samples that have most impact on the final results.

The network structure for stage-2 network is shown in Figure \ref{fig:stage2}. The visual feature for the input image is obtained by a visual CNN.
Word features are generated by the encoder LSTM. At each word, a joint word-image feature is obtained via the spatial attention module, which relates the word feature to its corresponding image regions.
A decoder LSTM then automatically aligns encoded features for the words to enhance robustness against sentence structure variations. The output features of the decoder LSTM is utilized to obtain the final matching confidence.
The idea of spatial and latent semantic co-attention was for the first time proposed and the network is accordingly designed. Unlike LSTM decoders for NLP \cite{bahdanau2014neural,vinyals2015show}, whose each step corresponds to a specific output word, each step of our semantic decoder captures a latent semantic concept and the number of steps is predefined as the number of concepts.

\subsubsection{Encoder word-LSTM with spatial attention}
For the visual CNN and encoder LSTM, our goal is to generate a joint word-visual feature representation at each input word. The naive solution would be simply concatenating the visual feature with word feature at each word.
However, different words or phrases may relate more to specific visual regions instead of the overall image. Inspired by \cite{vinyals2015show}, we adopt a spatial attention mechanism to weight more on relevant visual regions for each word.

Given an input sentence description, we first encode each word to an one-hot vector and then transform them to a feature vector through a fully-connected layer and an encoder word-LSTM. We denote the word features by $H={\{h_1, \cdots ,h_T\}}$, $H \in \mathbb{R}^{D_H \times T}$, where $h_t$ denotes the hidden state of the encoder LSTM at time step $t$ and $D_H$ is the hidden state dimension. Let $I = {\{i_1, \cdots ,i_L\}}$, $I \in \mathbb{R}^{D_I \times L}$ represent the visual features from all $L$ regions in the input image, where $D_I$ is the image feature dimension, and $i_l$ is the feature vector at the spatial region $l$. At time step $t$, the spatial attention $a_t$ over each image region $k$ can be calculated as
\begin{align}
e_{t,k} = &\,\,W_P\left\{ \tanh\left[W_I i_k + (W_H h_t + b_H) \right] \right\} + b_P,\\
a_{t,k} = & \,\, \frac{\exp(e_{t,k})}{\exp\left(\sum_{j=1}^L e_{t,j}\right)}, \textnormal{  for $k= 1, \cdots, L$},
\end{align}
where $W_I \in \mathbb{R}^{K\times D_I}$ and $W_H \in \mathbb{R}^{K \times D_H}$ are the parameter matrices that transform visual and semantic features to the same $K$-dimensional space, and $W_P \in \mathbb{R}^{1 \times K}$ converts the coupled textual and visual features to affinity scores. Given a word at time $t$, the attentions $a_{t,k}$ over all $L$ image regions are normalized by a softmax function and should sum up to 1.
Intuitively, $a_{t,k}$ represents the probability that the $t$th word relates to the $k$th image region.
The obtained spatial attentions are then used to gate the visual features by weighted averaging,
\begin{align}
\tilde{i}_t = \sum_{k=1}^{L}{a_{t,k} i_k}.
\end{align}
In this way, the gated visual features focus more on relevant regions to the $t$th word.
To incorporate both textual and visual information at each word, we then concatenate the gated visual features $\tilde{i}_t$ and hidden states $h_t$ of LSTM as the output of the spatial attention module, $x_t = \left[\tilde{i}_t, h_t\right]$.
\subsubsection{Decoder LSTM with latent semantic attention}
Although the LSTM has a memory state and a forget gate to capture long-term information, it still faces challenges on processing very long sentences to compress all information of the input sentence into a fixed-length vector. It might not be robust enough against sentence structure variations. Inspired by the word alignment (attention) technique in neural machine translation \cite{bahdanau2014neural}, we propose to use a decoder LSTM with latent semantic attention to automatically align sentence structures and estimate the final matching confidence. Note that unlike the conventional decoder LSTM in machine translation, where each step corresponds to an actual word, each step of our decoder LSTM has no physical meaning but only latent semantic meaning.
Given the final features encoded by the encoder LSTM, the $M$-step decoder LSTM processes the encoded feature step by step while searches through the entire input sentence to align the image-word features, $x_t, t=\{1, \cdots, T\}$.
At the $m$th time step of the decoding process, the latent semantic attention $a'_{m}$ for the $t$th input word is calculated as
\begin{align}
e'_{m, t} &= f(c_{m-1}, x_t), \\
a'_{m, t} &= \frac{\exp (e'_{m,t})} {\sum_{j=1}^{T} {\exp (e'_{m,j}) } },
\label{eq:3}
\end{align}
where $f$ is an importance function that weights the importance of the $j$th word for the $m$th decoding step. It is modeled a two-layer Convolutional Neural Network. $c_{m-1}$ is the hidden state by decoder LSTM for step $m-1$.
At each decoding step $m$, the semantic attention ``soft'' aligns the word-image features by a weighted summation,
\begin{align}
\tilde{x}_m = \sum_{j=1}^{T} {a'_{m,j} x_j}.
\end{align}
The aligned image-word features $\tilde{x}_m$ are then transformed by two fully-connected layers and fed into the $M$-step decoding LSTM to obtain the final matching confidence.
By automatically aligning image-word features with latent semantic attention, at each decoding step, the decoder LSTM is able to focus more on relevant information by re-weighting the source image-word features to enhance the network's robustness to sentence structure variations.
For training the stage-2 network, we also utilize identity-level annotations when constructing text-image training pairs. If an image and a sentence have the same identity, they are treated as a positive pair. Easier training samples are filtered out by the stage-1 network. The decoder LSTM is trained with the binary cross-entropy loss,
\begin{align}
  E = -\frac{1}{N'}\sum_{i=1}^{N'} \left[ y_i \log C_i  + (1-y_i) \log (1-C_i) \right]
\end{align}
where $N'$ is the number of samples for training the stage-2 network, $C_i$ is the predicted matching confidence for the $i$th text-image pair, and $y_i$ denotes its target label, with 1 representing the text and image pair belonging to the same identity and 0 representing different identities.

\vspace{-9pt}
\section{Experiments}
\subsection{Datasets and evaluation metrics}
Our proposed algorithm takes advantage of identity-level annotations from the data for achieving robust matching results. Three datasets with identity-level annotations, CUHK-­PEDES \cite{li2017person}, Caltech-UCSD birds (CUB) \cite{reed2016learning}, and Oxford-102 Flowers \cite{reed2016learning}, are chosen for evaluation.

\textbf{CUHK-­PEDES dataset.}
The CUHK-­PEDES dataset \cite{li2017person} contains 40,206 images of 13,003 person identities. Each image is described by two sentences.
There are 11,003 persons, 34,054 images and 68,108 sentence descriptions in the training set. The validation set and test set consist of 3,078 and 3,074 images, respectively, and both of them contain 1,000 persons. The top-1 and top-10 accuracies are chosen to evaluate the performance of person search with natural language description following \cite{li2017person}, which are the percentages of successful matchings between the query text and the top-1 and top-10 scored images. 

\textbf{CUB dataset and Flower dataset.}
The CUB and Flower datasets contain 11,788 bird images and 8,189 flower images respectively, where each image is labeled by ten textual descriptions. There are 200 different categories in CUB and the dataset is splited into 100 training, 50 validation, and 50 test categories. Flower has 102 flower classes and three subsets, including 62 classes for training, 20 for validation, and 20 for test.
We have the same experimental setup as \cite{reed2016learning} for fair comparison. There is no overlap between training and testing classes. Similar to \cite{reed2016learning}, identity classes are used only during training, and testing is on new identities.
We report the AP@50 for text-to-image retrieval and the top-1 accuracy for image-to-text retrieval. Given a query textual class, the algorithm first computes the percent of top-50 retrieved images whose identity matches that of the textual query class. The average matching percentage of all 50 test classes is denoted as AP@50. 

\begin{table}[]
\small
\centering
\begin{tabular}{|l|c|c|}
\hline
           & \multicolumn{2}{c|}{Text-Image Retrieval}                                     \\ \hline
Method                                          & Top-1 (\%)                              & Top-10 (\%)                       \\ \hline
deeper LSTM Q+norm I \cite{Antol_2015_ICCV}     & 17.19                                   & 57.82                                 \\ \hline
iBOWIMG \cite{zhou2015simple}                   & 8.00                                    & 30.56                                 \\ \hline
NeuralTalk \cite{vinyals2015show}               & 13.66                                   & 41.72                                 \\ \hline
Word CNN-RNN \cite{reed2016learning}            & 10.48                                   & 36.66                                 \\ \hline
GNA-RNN \cite{li2017person}                     & 19.05                                   & 53.64                                 \\ \hline
GMM+HGLMM \cite{klein2015associating}                     & 15.03                                   & 42.27                                 \\ \hline
\hline
Stage-1                                        & 21.55                                   & 54.78                                 \\ \hline
Stage-2                                      & {\color[HTML]{FE0000} \textbf{25.94}}                                   & {\color[HTML]{FE0000} \textbf{60.48}}                                 \\ \hline
\end{tabular}
\caption{Text-to-image retrieval results by different compared methods on the CUHK-­PEDES dataset \cite{li2017person}.}
\label{tab:reid_dataset}
\end{table}
\begin{table}[]
\small
\centering
\begin{tabular}{|l|c|c|}
\hline
                              & \multicolumn{2}{c|}{Text-Image Retrieval} \\ \hline
Method                        & Top-1 (\%)        & Top-10 (\%)              \\ \hline
Triplet      & 14.76             & 51.29              \\ \hline
Stage-1                       & 21.55             & 54.78              \\ \hline
Stage-2 w/o SMA+SPA+stage-1  & 17.19             & 57.82             \\ \hline
Stage-2 w/o SMA+SPA        & 22.11             & 58.05             \\ \hline
Stage-2 w/o SMA             & 23.58             & 58.68             \\ \hline
Stage-2 w/o ID              & 23.47             & 54.77             \\ \hline
Stage-2                       & {\color[HTML]{FE0000} \textbf{25.94}}    & {\color[HTML]{FE0000} \textbf{60.48}}   \\ \hline
\end{tabular}
\caption{Ablation studies on different components of the proposed two-stage framework. ``w/o ID'': not using identity-level annotations. ``w/o SMA'': not using semantic attention. ``w/o SPA'': not using spatial attention. ``w/o stage-1'': not using stage-1 network for training initialization and easy result screening.}
\vspace{-5pt}
\label{tab:ablation}
\end{table}

\subsection{Implementation details}
For fair comparison with existing baseline methods on different datasets, we choose VGG-16 \cite{simonyan2014very} for the CUHK-­PEDES dataset and GoogleNet \cite{szegedy2015going} for the CUB and Flower datasets as the visual CNN. For stage-1 network, the visual features are obtained by $L2$-normalizing the output features at ``drop7'' and ``avgpool'' layers of VGG-16 and GoogleNet.
We take the last hidden state of the LSTM to encode the whole sentence and embed the textual vector into the $512$-dimensional feature space with the visual image. The textual features is also $L2$-normalized.
The temperature parameters $\sigma_v$ and $\sigma_s$ in Eqs. (\ref{eq:1}) and (\ref{eq:2}) are empirically set to $0.04$. The LSTM is trained with the Adam optimizer with a learning rate of $0.0001$ while the CNN is trained with the batched Stochastic Gradient Descent.
For the stage-2 CNN-LSTM network, instead of embedding the visual images into 1-dimensional vectors, we take the output of the ``pool5'' layer of VGG-16 or the ``inception (5b)'' layer of GoogleNet as the image representations for learning spatial attention.
During the training phase, we first train the language model and fix the CNN model, and then fine-tune the whole network jointly to effectively couple the image and text features. The training and testing samples are screened by the matching results of stage-1. For each visual or textual sample, we take its $20$ most similar samples from the other modality by stage-1 network and construct textual-visual pair samples for stage-2 training and testing. Each text-image pair is assigned with a label, where 1 represents the corresponding pair and 0 represents the non-corresponding one. 
The step length $M$ of the decoding LSTM is set to $5$.

\begin{figure*}
\begin{center}
\includegraphics[width=1\linewidth]{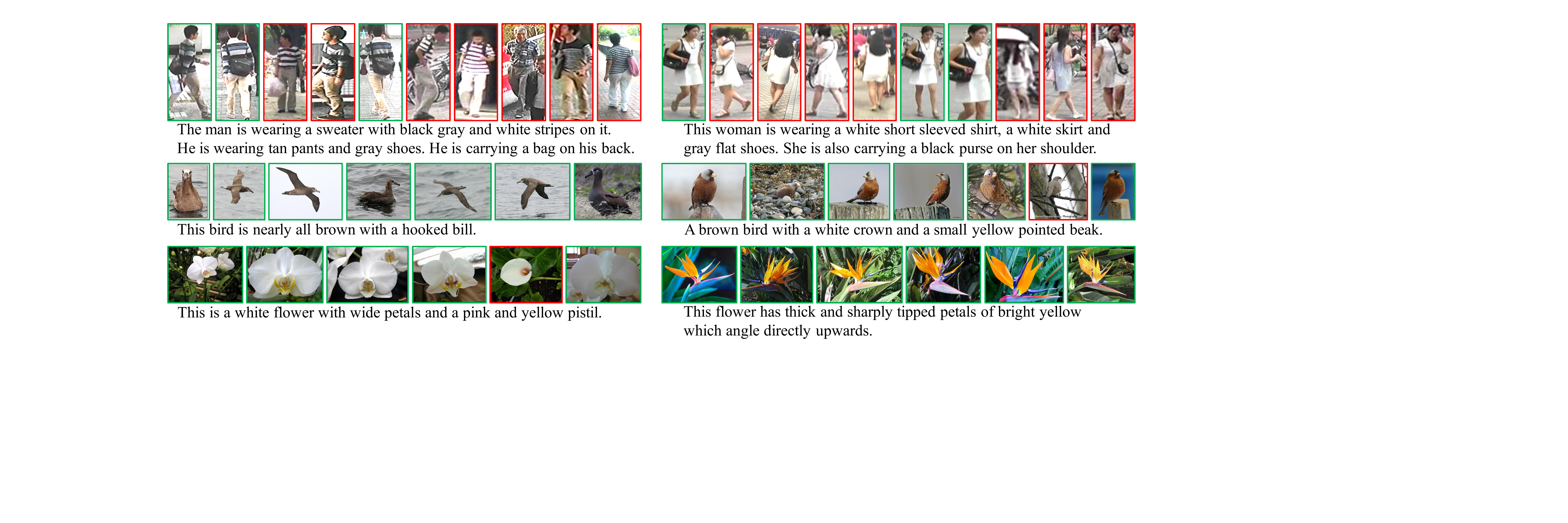} \ \\
\end{center}
\vspace{-11pt}
\caption{Example text-to-image retrieval results by the proposed framework. Corresponding images
are marked by green rectangles. (Left to right) For each text description, the matching results are sorted according to the similarity scores in a descending order. (Row 1) results from the CUHK-­PEDES dataset \cite{li2017person}. (Row 2) results from the CUB dataset \cite{reed2016learning}. (Row 3) results from the Flower dataset \cite{reed2016learning}.}
\vspace{-5pt}
\label{fig:retrieval}
\end{figure*}

\subsection{Results on CUHK-PEDES dataset}

We compare our proposed two-stage framework with six methods on the CUHK-­PEDES dataset. The top-1 and top-10 accuracies of text-to-image retrieval are recorded in Table \ref{tab:reid_dataset}. 
Note that only text-to-image retrieval results are evaluated for the dataset because image-to-text retrieval is not a practical problem setting for the dataset.
Our method outperforms state-of-the-art methods by large margins, which demonstrates the effectiveness of the proposed two-stage framework in matching textual and visual entities with identity-level annotations.

Our stage-1 model outperforms all the compared methods. The gain on top-1 accuracy by our proposed method is $2.50\%$ compared with the state-of-the-art GNA-RNN \cite{li2017person},
which has more complex network structure than ours. This shows the advantages of the CMCE loss.
Furthermore, the introduction of feature buffers make the comparison computation more efficient even with a large number of identities.
GMM+HGLMM \cite{klein2015associating} uses the Fisher Vector as a sentence representation by pooling the word2vec embedding of each word in the sentence.
The Word CNN-RNN \cite{reed2016learning} aims to minimize the distances between corresponding textual-visual pairs and maximize the distances between non-corresponding ones within each mini-batch. However, such a method is restricted by the mini-batch size and cannot be applied to dataset with a large number of identities. Our CMCE loss results in a top-1 accuracy of $21.55\%$, which outperforms the Word CNN-RNN's $10.48\%$.
The stage-1 CNN-LSTM with CMCE loss performs well on both accuracy and time complexity with its loosely coupled network structure.

The stage-2 CNN-LSTM with latent co-attention further improves the performances by $4.39\%$ and $5.70\%$ in terms of top-1 and top-10 accuracies.
The co-attention mechanism aligns visual regions with latent semantic concepts effectively to minimize the influence of sentences structure variations.
Compared with methods that randomly sample pairs, such as deeper LSTM Q+norm I \cite{Antol_2015_ICCV}, iBOWIMG \cite{zhou2015simple}, NeuralTalk \cite{vinyals2015show} and GNA-RNN \cite{li2017person}, our network focuses more on distinguishing the hard samples after filtering out most easy non-correponding samples by the stage-1 network.

\subsection{Ablation studies}
In this section, we investigate the effect of each component in the stage-1 and stage-2 networks by performing a series of ablation studies on the CUHK-­PEDES dataset.
We first investigate the importance of proposed CMCE loss.
We train our stage-1 model with the proposed loss replaced by triplet loss \cite{reed2016learning}, named ``Triplet''. As shown in Table \ref{tab:ablation}, its top-1 drops by 6.79$\%$ on the CUHK-PEDES set compared with our stage-1 with the new loss function. In addition, triplet loss \cite{reed2016learning} needs 3 times more training time.
Then we investigate the importance of the identity-level annotations to the textual-visual matching performance by ignoring the annotations. In this case, each image or sentence is treated as an independent identity. The top-1 and top-10 accuracies of ``Stage-2 w/o ID'' have $2.47\%$ and $5.71\%$ drops compared with the results of ``Stage-2'', which demonstrates that the identity-level annotations can help textual-visual matching by minimizing the intra-identity feature variations.

To demonstrate the effectiveness of our latent semantic attention, we remove it from the original stage-2 network, denoted as ``Stage-2 w/o SMA''. The top-1 accuracy drops by $2.36\%$, which shows the latent semantic attention can help align the visual and semantic concepts and mitigate the LSTM's sensitivity to different sentence structures.
The spatial attention tries to relate words or phrases to different visual regions instead of the whole image. Based on the framework of ``Stage-2 w/o SMA'', we further remove the spatial attention module from the stage-2 network, denoted as ``Stage-2 w/o SMA+SPA'', which can be viewed as a simple concatenation of the visual and textual features from the CNN and LSTM, followed by two fully-connected layers for binary classification. Both the top-1 and top-10 accuracies decrease compared with ``Stage-2 w/o SMA''.

The stage-1 network is able to provide samples for the training and evaluation of stage-2 network, and also serves as the initial point for its training. To investigate the influence of stage-1 network, we design one additional baselines, denoted as ``Stage-2 w/o SMA+SPA+Stage-1''. This baseline is trained without using the stage-1 network. It shows an apparent performance drop compared with the ``Stage-2 w/o SMA+SPA'' baseline, which  demonstrates the necessity of the stage-1 network in our proposed framework.
Since stage-1 network chooses only $20$ most closest images of each query text for stage 2 during the evaluation phase, the effect of some components might not be apparent in terms of the top-10 accuracy.

\subsection{Results on the CUB and Flower datasets}

Tables \ref{tab:CUB_dataset} and \ref{tab:Flower_dataset} show the experimental results of image-to-text and text-to-image retrieval on the CUB and Flower datasets. We compare with state-of-the-art methods on the two datasets.
The CNN-RNN \cite{reed2016learning} learns a CNN-RNN textual encoder for sentence feature embedding and transforms both visual and textual features into the same embedding space.
Different text features are also combined with the CNN-RNN methods.
The Word2Vec \cite{mikolov2013distributed} averages the pre-trained word vector of each word in the sentence description to represent textual features. BoW \cite{harris1954distributional} is the output of an one-hot vector passing through a single layer linear projection. Attributes \cite{akata2015evaluation} maps attributes to the embedding space by learning a encoder function. Different types of textual representations are combined with the CNN-RNN framework for testing.
Our method outperforms the state-of-the-art CNN-RNN by more than $3\%$ in terms of top-1 image-to-text retrieval accuracy and about $10\%$ in terms of text-to-image retrieval AP@50 on both datasets, which shows the effectiveness of the proposed method.
For the ``Triplet'' baseline, the top-1 and AP@50 drop by 9.0$\%$ and 3.1$\%$ on CUB dataset, and drop by 4.1$\%$ and 3.1$\%$ on Flower dataset which demonstrate the proposed loss function performs better than the traditional triplet loss.
Since the top-1 accuracy provided by \cite{reed2016learning} is computed by fusing sentences of the same class into one vector and our stage-2 network is therefore not suitable for the image-to-text retrieval task, we only report the stage-1 results on image-to-text retrieval which has already outperformed other baselines.

\setlength{\tabcolsep}{3pt}
\begin{table}[]
\small
\centering
\begin{tabular}{|l|c|c|c|c|}
\hline
             & \multicolumn{2}{c|}{Image-Text} & \multicolumn{2}{c|}{Text-Image}                           \\ \hline
             & \multicolumn{2}{c|}{Top-1 Acc (\%)} & \multicolumn{2}{c|}{AP@50 (\%)}                           \\ \hline
Methods      & DA-SJE        & DS-SJE              & DA-SJE                      & DS-SJE                      \\ \hline
BoW \cite{harris1954distributional} & 43.4          & 44.1                & 24.6                        & 39.6                        \\ \hline
Word2Vec \cite{mikolov2013distributed}     & 38.7          & 38.6                & 7.5                         & 33.5                        \\ \hline
Attributes \cite{akata2015evaluation}   & 50.9          & 50.4                & 20.4                        & 50.0                        \\ \hline
Word CNN \cite{reed2016learning}    & 50.5          & 51.0                & 3.4                         & 43.3                        \\ \hline
Word CNN-RNN \cite{reed2016learning}  & 54.3          & 56.8                & 4.8                         & 48.7                        \\ \hline
GMM+HGLMM \cite{klein2015associating} & \multicolumn{2}{c|}{36.5}           & \multicolumn{2}{c|}{35.6} \\ \hline
Triplet	 & \multicolumn{2}{c|}{52.5}           & \multicolumn{2}{c|}{52.4} \\ \hline
\hline
Stage-1      & \multicolumn{2}{c|}{61.5}           & \multicolumn{2}{c|}{55.5}                                 \\ \hline
Stage-2     & \multicolumn{2}{c|}{$-$}               & \multicolumn{2}{c|}{{\color[HTML]{FE0000} \textbf{57.6}}} \\ \hline
\end{tabular}
\caption{Image-to-text and text-to-image retrieval results by different compared methods on the CUB dataset \cite{reed2016learning}.}
\label{tab:CUB_dataset}
\end{table}

\begin{table}[]
\small
\centering
\begin{tabular}{|l|c|c|c|c|}
\hline
             & \multicolumn{2}{c|}{Image-Text} & \multicolumn{2}{c|}{Text-Image}                           \\ \hline
             & \multicolumn{2}{c|}{Top-1 Acc (\%)} & \multicolumn{2}{c|}{AP@50 (\%)}                           \\ \hline
Methods      & DA-SJE        & DS-SJE              & DA-SJE                      & DS-SJE                      \\ \hline
BoW \cite{harris1954distributional}       & 56.7          & 57.7                & 28.2                        & 57.3                        \\ \hline
Word2Vec \cite{mikolov2013distributed}     & 54.6          & 54.2                & 16.3                        & 52.1                        \\ \hline
Word CNN \cite{reed2016learning}    & 60.2          & 60.7                & 8.7                         & 56.3                        \\ \hline
Word CNN-RNN \cite{reed2016learning} & 60.9          & 65.6                & 7.6                         & 59.6                        \\ \hline
GMM+HGLMM \cite{klein2015associating} & \multicolumn{2}{c|}{54.8}           & \multicolumn{2}{c|}{52.8} \\ \hline
Triplet	 & \multicolumn{2}{c|}{64.3}           & \multicolumn{2}{c|}{64.9} \\ \hline
\hline
Stage-1      & \multicolumn{2}{c|}{68.4}           & \multicolumn{2}{c|}{68.0}                                 \\ \hline
Stage-2      & \multicolumn{2}{c|}{$-$}               & \multicolumn{2}{c|}{{\color[HTML]{FE0000} \textbf{70.1}}} \\ \hline
\end{tabular}
\caption{Image-to-text and text-to-image retrieval results by different compared methods on the Flower dataset \cite{reed2016learning}.}
\label{tab:Flower_dataset}
\end{table}

\subsection{Qualitative results}
We also conduct qualitative evaluations of the proposed methods. Figure \ref{fig:retrieval} shows example text-to-image retrieval results. Most sentences can correctly match images corresponding to their descriptions. In the first case, almost all the persons wear a sweater with ``black gray and white stripes''. Different images of the same identity (the first, second, and fifth person images) appear in the top-ranked results, which shows the proposed two-stage CNN-LSTM can correctly match identities across different domains and minimizes the intra-identity distances. Some mis-matching results are even challenging for human to distinguish with subtle differences in visual appearance. In the second case, the first and second person both wear ``white short sleeved shirt'', but only the first one is the true matching result because of the ``black purse'' carried on her shoulder.
\section{Conclusion}
In this paper, we proposed a novel two-stage framework for identity-aware visual-semantic matching. The framework consists of two deep neural networks. The stage-1 CNN-LSTM network learns to embed the input image and description to the same feature space and minimizes the intra-identity distance simultaneously with the CMCE loss. It serves as initial point for stage-2 training and also provides training and evaluation samples for stage-2 by screening most incorrect matchings.
The stage-2 network is a CNN-LSTM with latent co-attention mechanism which jointly learns the spatial attention and latent semantic attention by an alignment decoder LSTM. It automatically aligns different words and image regions to minimize the impact of sentence structure variations. We evaluate the proposed method on three datasets and perform a series of ablation studies to verify the effect of each component. Our method outperforms state-of-the-art approaches by a large margin and demonstrates the effectiveness of the proposed framework for identity-aware visual-textual matching.

\vspace{5pt}
\noindent
{\bf Acknowledgement}
This work is supported in part by SenseTime Group Limited, in part by the General Research Fund through the Research Grants Council of Hong Kong under Grants CUHK14213616, CUHK14206114, CUHK14205615, CUHK419412, CUHK14203015, CUHK14239816, CUHK14207814, in part by the Hong Kong Innovation and Technology Support Programme Grant ITS/121/15FX, and in part by the China Postdoctoral Science Foundation under Grant 2014M552339.

{\small
\bibliographystyle{ieee}
\bibliography{egbib}
}

\end{document}